\documentclass[conference]{IEEEtran}

\usepackage{mytexpckgs}

\begin{document}
%
\title{Modular Deep Recurrent Neural Network: Application to Quadrotors}


\author{\IEEEauthorblockN{Nima Mohajerin}
\IEEEauthorblockA{Dept. of Mechanical and Mechatronics Engineering\\
University of Waterloo, 
Waterloo, ON,Canada\\
nima.mohajerin@uwaterloo.ca}
\and
\IEEEauthorblockN{Steven L. Waslander}
\IEEEauthorblockA{Dept. of Mechanical and Mechatronics Engineering\\
University of Waterloo, 
Waterloo, ON, Canada\\
stevenw@uwaterloo.ca}}

\graphicspath{ {./Figures/} }

\maketitle
\maketitle

\begin{center}
	\footnotesize
	\textit{This paper was originally published in the Proceedings of the
		2014 IEEE International Conference on Systems, Man, and Cybernetics (SMC),
		San Diego, CA, USA, 2014. The published version is available at
		\href{https://doi.org/10.1109/SMC.2014.6974106}{https://doi.org/10.1109/SMC.2014.6974106}.}
\end{center}

\begin{abstract}
A modular deep Recurrent Neural Network (RNN) is introduced to facilitate the process of deploying various architectures of RNNs, and to automatically compute derivatives for gradient-based learning methods. The modularity leads to a set of new architectures, one of which includes feedforward inter-layer connections.  By adding feedforward inter-layer connections in a multi-layer RNN, it is observed that the capability of the RNN to learn and model high-order dynamics and nonlinearities is significantly improved. The problem of vanishing/exploding gradient in space for a multilayer RNN is also alleviated using feedforward connections. These results are demonstrated using a quadrotor case study, for which a model of the altitude dynamics is learned with our particular network structure, while existing methods are unable to generalize as quickly or at all.
\end{abstract}
\IEEEpeerreviewmaketitle

\section{Introduction}
Recurrent Neural Networks have been proposed for modelling dynamical systems using empirical data~\cite{Narendra90,Tsung1994}. Among various architectures, Nonlinear AutoRegresive with eXogenous input models (NARX) (and similar architectures, such as Nonlinear AutoRegresive Moving Average with eXogenous inputs - NARMAX) and  Recurrent MultiLayer Perceptron (RMLP) have received a great amount of attention~\cite{Kumar2006,Perez2013,Wang06}. It has been shown that NARX is capable of learning dynamical systems~\cite{Lin1996:NARX}. In~\cite{Kumar2006}, the authors compare NARX with RMLP and a Memory Neuron Network (MNN) for identification of helicopter dynamics, and demonstrate that NARX is the most suitable method. However, the selection of number of delays poses a particular challenge in implementation~\cite{Delgado95},  as a fairly good knowledge of the system to be modeled is required \textit{a priori} in order to define sufficient delay units on a NARX input and output without over-specifying the network~\cite{Kumar2006}.

Other RNN architectures have also been explored for dynamic system modelling~\cite{Deng2013,Wang06,Nouri2002,Perez2013}. Most of the architectures, however, are made of at most two hidden layers and feedback is used both inter and intra-layer. Although adding more layers to the RNN architecture will increase the capacity of the network to represent higher order and more complex nonlinearities, as in deep neural networks~\cite{Hinton2006:1,Hinton2006:2}, usually using multiple layers is avoided because of the so-called problem of \emph{vanishing/exploding gradient}~\cite{Bengio94}. In fact, one of the main reasons to prevent the application of gradient-descent optimization methods to train RNNs is that as the error gradient is propagated both back in time and towards former layers inside the network, it either explodes or vanishes, leading to an ill-conditioned Jacobian and numerical instability in the optimization. This problem exists both in the time and space domains of an RNN.

The motivation for our work stems from the need to identify accurate models for small unmanned aerial vehicles (UAVs), such as quadrotor helicopters.  In recent years, quadrotors use has grown rapidly, both in research and industry, due to their ease of use, low maintenance requirements and agility of motion.  Additionally, quadrotors possess interesting and complex dynamics which remain challenging to model accurately through all phases of flight. More accurate models are essential to enabling more precise flight control, which in turn will allow quadrotors to be used closer to objects in the environment and at higher speeds.

There have been quite a few successful attempts to capture the main dynamics of the quadrotor using physical properties of the system~\cite{hoffmann2011,madani2008}. However, modelling all aerodynamic effects, sensor and actuator uncertainties and all other nonlinearities using pure physical attributes of the vehicle  is a very difficult task and requires wind tunnel testing to accurately assess aerodynamic coefficients. Therefore, a modelling approach based on empirical data collected from typical flight experiments may significantly reduce the effort required to define accurate models of quadrotor vehicles. Because quadrotors are dynamic systems, clearly this approach needs to be capable of capturing both dynamic characteristics and nonlinearities of the vehicle model. 

In this paper, we aim at addressing the problem of vanishing/exploding gradient (in space) to add multiple layers to RNNs and make them capable of modelling high-order dynamical nonlinear systems such as a quadrotor. We will formulate a MOdular DEep Recurrent Neural Network (MODERNN) architecture which covers a variety of recurrent architectures and facilitates the derivative calculation process. While this architecture can have multiple layers (hence the name deep), we will demonstrate that the presence of feedforward connections throughout a MODERNN contributes to decreasing the vanishing/exploding gradient problem in space, or equivalently the network output Jacobian will not be ill-conditioned even with multiple layers. Consequently, ordinary gradient descent training methods, such as the Levenberg-Marquardt algorithm (LM) are still applicable to train MODERNN. Through comparison with NARX and RMLP, we will assess the performance of MODERNN on modelling altitude of a quadrotor.

This work proceeds as follows. In the next section (Section~\ref{sec:arch}), our proposed RNN architecture is presented in detail. A modular method to obtain the derivatives of the network output(s) with respect to the network weights is presented afterwards in Section~\ref{sec:deriv}. Then the learning algorithm for training MODERNN is described in Section~\ref{sec:learn}. In Section~\ref{sec:sim}, the simulation results of MODERNN modelling the altitude of a quadrotor are presented and compared with NARX and regular Multilayer RNN (MRNN). Finally, Section~\ref{sec:conc} will conclude the paper and outline future work.	
\vspace{-5pt}
\section{Modular Deep Recurrent Neural Network}
\label{sec:arch}
Our network definition is a generalization over RMLP with locally recurrent layers ($G_i$) and connections between all layers. An example of a three layer MODERNN is illustrated in Fig.~\ref{fig:MODERNN:EX1} . Each block $G_i$ represents a locally recurrent layer. In fact, $G_i$ is a dynamic MIMO system with $m_i$ inputs and $n_i$ outputs. The equations governing the dynamic of $G_i$ are as follows:
\vspace{-5pt}
\begin{equation}
	\left\{ 
  		\begin{array}{l l}
			\mathbf{x}_i(k) = \mathbf{A}_i\mathbf{y}_i(k-1) + \mathbf{B}_i\mathbf{u}_i(k)+\mathbf{b}_i\\
   			\mathbf{y}_i(k) = \mathbf{f}_i\big{(}\mathbf{x}_i(k)\big{)},
  \end{array} \right.
\label{eq:LRLayer}
\end{equation}
where $k$ is the discrete-time index, and:
\begin{itemize}
	\item $\mathbf{x}_i(k)\in\mathbb{R}^{n_i}$ is the state of the layer,
	\item $\mathbf{y}_i(k)\in\mathbb{R}^{n_i}$ is the output of the layer,
	\item $\mathbf{u}_i(k)\in\mathbb{R}^{m_i}$ is the input to the layer,
	\item $\mathbf{A}_i\in\mathbb{R}^{n_i}\times\mathbb{R}^{n_i}$ is the feedback weight matrix,
	\item $\mathbf{B}_i\in\mathbb{R}^{n_i}\times\mathbb{R}^{m_i}$ is the input weight matrix,
	\item $\mathbf{b}_i\in\mathbb{R}^{n_i}$ is a bias weight vector,
	\item $\mathbf{f}_i(.)$ is the layer activation function,
	\item $n_i$ is the number of the neurons inside the layer,
	\item $m_i$ is the number of input signals to the layer.
\end{itemize}

We will refer to $\mathbf{x}_i(k)$ and $\mathbf{y}_i(k)$ as the network states as long as $\mathbf{y}_i(k)$ does not form the network output(s). Also, we will refer to the vectorized form of the weights inside the layer $G_i$ by $\mathbf{p}_i$:
\begin{equation}
	\begin{aligned}
		\mathbf{p}_i^T=[ \mathbf{B}_i(:,1)^T \hspace{5pt}& \dots& \hspace{1pt}  &\mathbf{B}_i(:,m_i)^T \\
					     \mathbf{A}_i(:,1)^T  \hspace{5pt}& \dots &\hspace{1pt}  &\mathbf{A}_i(:,n_i)^T \hspace{5pt}  \mathbf{b}_i^T],
	\end{aligned}
	\vspace{-5pt}
\end{equation}
where $ \mathbf{A}_i(:,j)^T$ is the transpose of the $j$\textsuperscript{th} column of $\mathbf{A}_i$ and similarly for  $\mathbf{B}_i$. The vector $\mathbf{p}_i\in\mathbb{R}^{q_i}$, where $q_i=n_i(m_i+n_i+1)$ is the number of all weights inside $G_i$. Note that $\mathbf{u}_i(k)$ includes the independent inputs (i.e. $\mathbf{u}(k)$) as well as outputs from other layers ($\mathbf{y}_j(k-1), j\neq i$). Note also that the output of a layer is actually the output of all neurons inside the layer.

Although we can include $\mathbf{y}_i(k-1)$ inside $\mathbf{u}_i(k)$ and make the equations more compact, we avoid this for two reasons. First, our proposed formulation is more modular, hence, easier to implement. Second, the process of calculating the derivatives is more straightforward, as will be seen shortly.

With inclusion of a \emph{connection matrix} we can modify the inter-layer connection in a MODERNN. The proposed connection matrix, $\mathbf{C}$, is an $(L+2)\times L$ matrix for an $L$ layer network. Elements of $\mathbf{C}$, i.e. $c_{ij}$, represent absence/presence of a connection between layer $i$ and $j$, by being either 0 or 1, where row $L+1$ corresponds to connections from the independent input signals ($\mathbf{u}(k)$) and row $L+2$ corresponds to connections forming the network output signal. For example, the connection matrix for the example shown in Fig.~\ref{fig:MODERNN:EX1}, with fully connected sublayers, feedforward to each layer and network output dependent on only the last layer's outputs, is given by,
\vspace{-5pt}
\begin{equation}
	\small
	\mathbf{C}=\begin{bmatrix}
		1 & 1 & 1 \\
		1 & 1 & 1 \\
		1 & 1 & 1 \\
		1 & 1 & 1\\
		0 & 0 & 1
	\end{bmatrix}.
\end{equation}
Using this connection matrix, various architectures are attainable. For instance, to have an RMLP with 4 layers, the following connection matrix is used,
\begin{equation}
	\small
	\mathbf{C}=\begin{bmatrix}
		1 & 1 & 0 & 0 \\
		0 & 1 & 1 & 0 \\
		0 & 0 & 1 & 1 \\
		0 & 0 & 0 & 1 \\
		1 & 0 & 0 & 0 \\
		0 & 0 & 0 & 1
	\end{bmatrix}.
	\label{eq:CRMLP}
\end{equation}
Note that in an RMLP the layers are locally recurrent and sequentially connected, i.e. each layer is only connected to itself and the next layer. Therefore, in equation~\eqref{eq:CRMLP}, $c_{i,i}=1$ for $i=1,2,3,4$ correspond to local feedback and $c_{i,i+1}=1$  for $i=1,2,3$ correspond to the sequential connection. Also, $c_{4,1}=1$ means that the independent input is only connected to the first layer and $c_{6,1}=1$ means that the network output is the output from the final layer only.

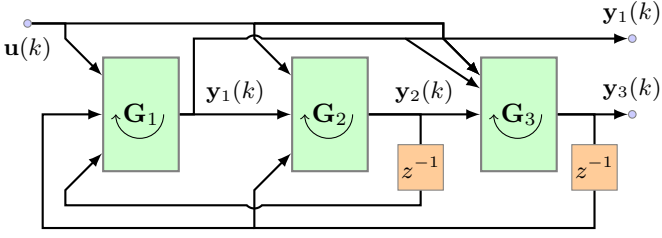
\begin{figure}
	\vspace{-5pt}
	\begin{tikzpicture}
		\node[NN,minimum height=1.5cm,minimum width=1cm] (G1) at (-2.5,0) []{$\mathbf{G}_1$};
		\draw [->] (-2.2,0) arc (0:-180:9pt);
		\node[NN,minimum height=1.5cm,minimum width=1cm] (G2) at (0.0,0) []{$\mathbf{G}_2$};
		\draw [->] (0.3,0) arc (0:-180:9pt);
		\node[NN,minimum height=1.5cm,minimum width=1cm] (G3) at (2.5,0) []{$\mathbf{G}_3$};
		\draw [->] (2.8,0) arc (0:-180:9pt);
		\node[sig] (u_k) at (-4,1.2)    [label=below:\small{$\mathbf{u}(k)$}] {};
		\node[sig] (y_k1) at (4,0)    [label=above:\small{$\mathbf{y}_3(k)$}] {};
		\node[sig] (y_k2) at (4,1)    [label=above:\small{$\mathbf{y}_1(k)$}] {};
		\node[delay] (z1) at (1.2,-0.7)  {\small{$z^{-1}$}};
		\node[delay] (z2) at (3.5,-0.7)  {\small{$z^{-1}$}};
		\draw[->,>=latex,thick] (u_k) to (-3.5,1.2) to (-3.5,1) to(G1);
		\draw[->,>=latex,thick] (u_k) to (-1,1.2) to (-1,1) to(G2);
		\draw[->,>=latex,thick] (-1,1.2) to (1.5,1.2) to (1.5,1) to(G3);
		\draw[->,>=latex,thick] (G1) -- (G2) node [above,text width=3cm,text centered,midway] {\small{$\mathbf{y}_1(k)$}};
		\draw[->,>=latex,thick] (G2) -- (G3) node [above,text width=3cm,text centered,midway] {\small{$\mathbf{y}_2(k)$}};
		\draw[->,>=latex,thick] (G3) -- (y_k1) ;
		\draw[->,>=latex,thick] (-1,1.2) to (1.5,1.2) to (1.5,1) to(G3);
		\draw[->,>=latex,thick] (G1) to (-1.8,0) to (-1.8,1.0) to (-1.1,1.0) to [rounded corners] (-1,1.1) to (-0.9,1) to (1,1) to (1,1) to(G3);
		\draw[->,>=latex,thick] (1,1.0) to (y_k2);
		\draw[->,>=latex,thick] (G2) to (1.2,0) to (z1) to (1.2,-1.2) to (-0.9,-1.2) to [rounded corners] (-1.0,-1.3) to (-1.1,-1.2) to  (-3.5,-1.2) to (-3.5, -1)  to (G1);
		\draw[->,>=latex,thick] (G3) to (3.5,0) to (z2) to (3.5,-1.5) to (-1,-1.5) to (-1,-1) to (G2);
		\draw[->,>=latex,thick] (-1,-1.5) to (-3.8,-1.5) to (-3.8,0) to(G1);
	\end{tikzpicture}
	\caption{A 3 layer MODERNN,with output $\mathbf{y}(k)$ equal to $\Big{[}\mathbf{y}_1^T(k) \hspace{5pt} \mathbf{y}_3^T(k)$\Big{]}.}
	\label{fig:MODERNN_middle_output}
\vspace{-25pt}
\end{figure}

Other forms of architectures are also attainable. For example, the architecture shown in Fig.~\ref{fig:MODERNN_middle_output} is a MODERNN in which the outputs are derived from the first and last layers. The connection matrix corresponding to this architecture is,
\begin{equation}
	\small
	\mathbf{C}=\begin{bmatrix}
		1 & 1 & 1 \\
		1 & 1 & 1 \\
		1 & 1 & 1 \\
		1 & 1 & 1\\
		1 & 0 & 1
	\end{bmatrix}.
	\vspace{-9pt}
\end{equation}
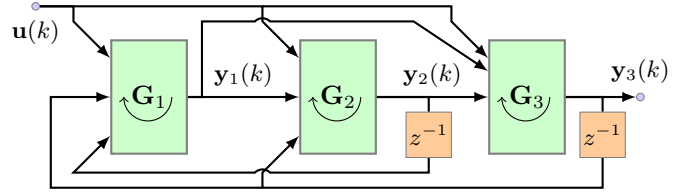
\begin{figure}
	\begin{tikzpicture}
		\node[NN,minimum height=1.5cm,minimum width=1cm] (G1) at (-2.5,0) []{$\mathbf{G}_1$};
		\draw [->] (-2.2,0) arc (0:-180:9pt);
		\node[NN,minimum height=1.5cm,minimum width=1cm] (G2) at (0.0,0) []{$\mathbf{G}_2$};
		\draw [->] (0.3,0) arc (0:-180:9pt);
		\node[NN,minimum height=1.5cm,minimum width=1cm] (G3) at (2.5,0) []{$\mathbf{G}_3$};
		\draw [->] (2.8,0) arc (0:-180:9pt);
		\node[sig] (u_k) at (-4,1.2)    [label=below:\small{$\mathbf{u}(k)$}] {};
		\node[sig] (y_k) at (4,0)    [label=above:\small{$\mathbf{y}_3(k)$}] {};
		\node[delay] (z1) at (1.2,-0.5)  {\small{$z^{-1}$}};
		\node[delay] (z2) at (3.5,-0.5)  {\small{$z^{-1}$}};
		\draw[->,>=latex,thick] (u_k) to (-3.5,1.2) to (-3.5,1) to(G1);
		\draw[->,>=latex,thick] (u_k) to (-1,1.2) to (-1,1) to(G2);
		\draw[->,>=latex,thick] (-1,1.2) to (1.5,1.2) to (1.5,1) to(G3);
		\draw[->,>=latex,thick] (G1) -- (G2) node [above,text width=3cm,text centered,midway] {\small{$\mathbf{y}_1(k)$}};
		\draw[->,>=latex,thick] (G2) -- (G3) node [above,text width=3cm,text centered,midway] {\small{$\mathbf{y}_2(k)$}};
		\draw[->,>=latex,thick] (G3) -- (y_k) ;
		\draw[->,>=latex,thick] (G1) to (-1.8,0) to (-1.8,1) to (-1.1,1) to [rounded corners] (-1,1.1) to (-0.9,1) to (1,1) to (1,1) to(G3);
		\draw[->,>=latex,thick] (G2) to (1.2,0) to (z1) to (1.2,-1.0) to (-0.9,-1.0) to [rounded corners] (-1.0,-0.9) to (-1.1,-1.0) to  (-3.5,-1.0) to (-3.5, -1)  to (G1);
		\draw[->,>=latex,thick] (G3) to (3.5,0) to (z2) to (3.5,-1.2) to (-1,-1.2) to (-1,-1) to (G2);
		\draw[->,>=latex,thick] (-1,-1.2) to (-3.8,-1.2) to (-3.8,0) to(G1);
	\end{tikzpicture}
	\caption{A 3 layer MODERNN, with output $\mathbf{y}(k)$ equal to $\mathbf{y}_3(k)$.}
	\label{fig:MODERNN:EX1}
\vspace{-25pt}
\end{figure}

As derivatives are required for each network definition to employ gradient-based learning methods, we present a generalized output Jacobian calculation, for all architectures derived from MODERNN using a connection matrix similar to what we already presented. However, we will use MODERNN to refer to a network with all inter-layer connections present, because we will observe that the presence of inter-layer connections is a crucial fact to enhance the network training process. Such a network with $L$ layers is presented as a function that maps inputs ($\mathbf{u}(k)\in\mathbb{R}^m$) to outputs ($\mathbf{y}(k)\in\mathbb{R}^n$):
\begin{equation}
	\mathbf{y}(k)=\Omega_{L}(\mathbf{u}(k)).
	\label{eq:MODERNN}
\end{equation}
For a more compact formulation, assume the input sequence is defined over a time horizon, $T$, i.e., $k = {1, \ldots, T}$. Then, we can arrange $\mathbf{u}(k)$ (for all $k=1,\dots,T$) inside a matrix $\mathbf{U}\in\mathbb{R}^m\times\mathbb{R}^T$ and pass it to the network.
\begin{equation}
	\mathbf{Y}=\Omega_{L}(\mathbf{U}).
	\label{eq:MODERNN_ALL}
\end{equation}
The output becomes a matrix $\mathbf{Y}\in\mathbb{R}^n\times\mathbb{R}^T$ where the $k$\textsuperscript{th} column of $\mathbf{Y}$, i.e. $\mathbf{y}(k)$ is calculated by equation~\eqref{eq:MODERNN} .
\section{Network Jacobians for MODERNN}
\label{sec:deriv}
In this section, we present a modular method to derive the network Jacobians. It has been reported that for different architectures, obtaining network output derivatives is a time consuming process~\cite{Martens2012}.  We attempt to address this problem and speed up the process of modifying/deploying present/new architectures.

To calculate the network Jacobians, we form $\mathbf{p}\in\mathbb{R}^q$ which is a vector encompassing all the weights inside a MODERNN network, i.e.,
\vspace{-10pt}
\begin{equation}
	\vspace{-5pt}
	\mathbf{p}^T=\begin{bmatrix} \mathbf{p}_1^T & \dots & \mathbf{p}_L^T \end{bmatrix},q=\sum_{i=1}^{L}q_i,
\end{equation}
where $L$ is the number of locally recurrent layers inside the network whose indexes are sorted in ascending order from input towards output. Additionally, the number of all weights inside this networks is given by $q$. 

Now, considering an arbitrary layer, $G_i$, we can write the Jacobian update rule as follows:
\begin{equation}
	\begin{aligned}
		\mathbf{J}_i^y(k)&=\frac{\partial\mathbf{y}_i(k)}{\partial\mathbf{p}}\\
		&= \mbox{diag}\Big{[}\mathbf{f}_i'\big{(}\mathbf{x}_i(k)\big{)}\Big{]}\Big{(}\mathbf{A}_i\mathbf{J}_i^y(k-1) + \mathbf{B}_i\mathbf{J}_i^u(k)+\Gamma_i(k)\Big{)}.
	\end{aligned}
	\label{eq:LRLayer_J}
\end{equation}
In equation~\eqref{eq:LRLayer_J}, $\mathbf{J}_i^y(k)\in\mathbb{R}^{n_i}\times\mathbb{R}^q$ is the Jacobian of the outputs of the layer $G_i$ at time $k$ and $\mathbf{J}_i^u(k)\in\mathbb{R}^{m_i}\times\mathbb{R}^q$ is the Jacobian of the inputs to that layer. $\Gamma_i(k)$ is a matrix which corresponds to the derivatives of the layer weights with respect to the network weights,
\begin{equation}
	\vspace{-5pt}
	\Gamma_i(k)=\frac{\partial\mathbf{A}_i}{\partial\mathbf{p}}\mathbf{y}_i(k-1) + \frac{\partial\mathbf{B}_i}{\partial\mathbf{p}}\mathbf{u}_i(k) + \frac{\partial\mathbf{b}_i}{\partial\mathbf{p}}.
\end{equation} 

Note that equation~\eqref{eq:LRLayer_J} is a recursive update for $\mathbf{J}_i^y(k)$. Therefore, we need to have an initial value for it at time $k=0$. Usually, because training is started from a stationary point, we assume zero initial condition for this matrix. Then, we need to form two matrices: $\mathbf{J}_i^u(k)$ and $\Gamma_i(k)$.  To form $\mathbf{J}_i^u(k)$, remember that
\begin{equation}
	\begin{aligned}
		\mathbf{u}_i(k)=\Big{[}& \mathbf{u}^T(k) \hspace{10pt} \mathbf{y}_1^T(k-1) \hspace{5pt} \dots \hspace{5pt} \mathbf{y}_{i-1}^T(k-1) \\ &\mathbf{y}_{i+1}^T(k-1)  \hspace{5pt} \dots \hspace{5pt} \mathbf{y}_L^T(k-1) \Big{]}^T.
	\end{aligned}
\end{equation}
Therefore, we see that
\vspace{-5pt}
\begin{equation}
	m_i=m+\sum_{\substack{j=1\\ j\neq i}}^Ln_j, \hspace{5pt}.
	\vspace{-5pt}
\end{equation}
Because the independent inputs are not dependent on the network weights, the first $m$ rows of $\mathbf{J}_i^u(k)$ are zero. 
\begin{equation}
	\begin{aligned}
	\mathbf{J}_i^u(k)=
		\Big{[}
			&\mathbf{0}^T_{m\times q} \hspace{5pt} {\mathbf{J}_1^y}^T(k-1)\hspace{5pt} \dots\hspace{5pt}  {\mathbf{J}_{i-1}^y}^T(k-1)\\
			&{\mathbf{J}_{i+1}^y}^T(k-1)\hspace{5pt}	\dots \hspace{5pt} {\mathbf{J}_L^y}^T(k-1)	
			\Big{]}^T.
	\end{aligned}
\end{equation}
The remaining elements inside the Jacobian $\mathbf{J}_i^u(k)$  are all calculated at the previous time step, and so this term can be determined recursively.

To modularly formulate $\Gamma_i(k)$, first let us define another matrix $\Lambda_i(k)$ for each layer, $G_i$, as follows:
\begin{equation}
	\Lambda_i(k)=\frac{\partial\mathbf{A}_i}{\partial\mathbf{p}_i}\mathbf{y}_i(k-1) + \frac{\partial\mathbf{B}_i}{\partial\mathbf{p}_i}\mathbf{u}_i(k) + \frac{\partial\mathbf{b}_i}{\partial\mathbf{p}_i}.
\end{equation}
In other words, $\Lambda_i(k)$ plays the same role as $\Gamma_i(k)$ but when the derivative is taken with respect to local weights only. Note, $\Lambda_i(k)$ is in $\mathbb{R}^{n_i}\times\mathbb{R}^{q_i}$. With some algebraic manipulation, the term $\Lambda_i(k)$ can be presented in the following form:
\begin{equation}
	\Lambda_i(k)=
		\begin{bmatrix}
			\Lambda_{i,1}(k) & \Lambda_{i,2}(k) &\mathbf{I}_{n_i}
		\end{bmatrix}_{n_i\times q_i},
		\vspace{-10pt}
\end{equation}
where
\begin{subequations}
	\begin{align}
		\Lambda_{i,1}(k)&=
			\begin{bmatrix}
				y_{i,1}(k-1)\mathbf{I}_{n_i} &  \dots & y_{i,n_i}(k-1)\mathbf{I}_{n_i} 
			\end{bmatrix}\\
		\Lambda_{i,2}(k)&=
			\begin{bmatrix}
				u_{i,1}(k)\mathbf{I}_{n_i} & \dots & u_{i,m_i}(k)\mathbf{I}_{n_i} 
			\end{bmatrix},
	\end{align}
\end{subequations}
and $u_{i,1}(k)$ refers to the first element inside $\mathbf{u}_i(k)$, and so on.

Now that we can calculate $\Lambda_i(k)$ for all layers, it is possible to form $\Gamma_i(k)$:
\begin{equation}
	\Gamma_i(k) = 
		\begin{bmatrix}
			\mathbf{0}_{n_i\times a} & \Lambda_i(k) & \mathbf{0}_{n_i\times b}
		\end{bmatrix}.
	\label{eq:Gamma-i}
\end{equation}
For equation~\eqref{eq:Gamma-i}, we define
\begin{equation}
	a=\sum_{j=1}^{i-1}q_j \hspace{5pt},  \hspace{10pt} 	b =\sum_{j=i+1}^Lq_j
\end{equation}
Now that $\Gamma_i(k)$ is defined, we can return to equation~\eqref{eq:LRLayer_J} and recursively update the Jacobian.

\section{A Learning Algorithm for Training MODERNN}
\label{sec:learn}

To address the problem of training RNN using network Jacobians, there are two commonly used methods: Back-Propagation Through Time (BPTT) and Real-Time Recurrent Learning (RTRL)~\cite{Pearlmutter,Pineda,Werbos90}. The time-series data which we aim to train the network with are trajectories in time. For our particular problem, i.e. modelling of a quadrotor's altitude dynamics, each training sample is a 2-tuple of a motor input (sum of motor speeds) and a position output (altitude of the vehicle), both over time, i.e. $\big{(}u_d(k),z_d(k)\big{)}$.  Because higher accuracy of the trained model is our final goal, we adopt an off-line, batch learning method to expose the network to more information. One of the most commonly used approaches to this end is the Least-Squares method, and the Levenberg-Marquardt algorithm (LM) is frequently reported to be very efficient in this context~\cite{Hagan94}. We employ LM to optimize a Sum-of-Squared Errors SSE cost function:
	\vspace{-5pt}
\begin{equation}
	E=0.5\sum_{k=1}^T\big{(}y(k)-z_d(k)\big{)}^2=0.5\mathbf{e}^T\mathbf{e},
	\vspace{-5pt}
\end{equation}
where:
\begin{equation}
	\vspace{-5pt}
	\begin{aligned}
		\mathbf{e}^T&=\begin{bmatrix} e(1) & e(2) & \dots & e(T)\end{bmatrix},\\
		e(k)&= y(k)-z_d(k),\text{ for }k=1,\dots,T.
	\end{aligned}
	\label{eq:ey}
	\vspace{-1pt}
\end{equation} 
Note that in our setup, the network has one output, which is the approximated altitude. We can have more than one, for example in case of a full quadrotor modelling, which involves more effort in computing the Jacobians.

\emph{Remark}: Because MODERNN has a recursive nature, it is necessary to initialize it properly. As the evolution of the network states is not known \textit{a priori} and is highly dependent on the network weights, initializing the untrained network to start from a non-zero initial condition is not reasonable as the network weights are merely random values at that point in the optimization. However, we propose that at the first stages of the training process, for each training sample, the network is initialized to zero. The training sample, i.e. the altitude envelop, should also start from a stationary point (e.g. quadrotor landed on the floor). The network weights should then converge to a set of more representative values. The training samples at these stages should encompass enough information from the beginning of a flight (e.g. both during take-off and with some variable altitude flight). Then, as the network is trained on these primary flight data, we can proceed to expanding the training horizon. For this work, because we are more interested in the MODERNN architecture and its capabilities, we focus on training the network with zeroed initial conditions.\hfill\(\Box\)

LM is essentially a second-order optimization method with variable step size which approximates the Hessian of the cost function with $\mathbf{J}^T\mathbf{J}$, where $\mathbf{J}$ is the Jacobian of the cost function~\cite{Levenberg}.  At each training iteration $k'$, the update rule in LM is given by,
\begin{equation}
	\Delta\mathbf{p}(k') = -(\mathbf{J}^T\mathbf{J}+\lambda\mathbf{I})^{-1}\mathbf{J}^T\mathbf{e}.
	\label{eq:LM}
\end{equation}
In equation~\eqref{eq:LM}, $\mathbf{J}$ is the Jacobian of the cost function over time given by,
\begin{equation}
	\mathbf{J}=
		\begin{bmatrix}
			\frac{\partial e(1)}{\partial p_1} & \dots & \frac{\partial e(1)}{\partial p_q} \\
			\vdots & \vdots & \vdots\\
			\frac{\partial e(T)}{\partial p_1} & \dots & \frac{\partial e(T)}{\partial p_q}
		\end{bmatrix}_{T\times q} = 
		\begin{bmatrix}
			\big{(}\frac{\partial e(1)}{\partial\mathbf{p}}\big{)}^T\\
			\vdots\\
			\big{(}\frac{\partial e(T)}{\partial\mathbf{p}}\big{)}^T
		\end{bmatrix}
		=\frac{\partial\mathbf{e}}{\partial\mathbf{p}}
	\label{eq:LMJ}
\end{equation}
\indent 
We propose a training method that divides the optimization into four nested loops. The outer most loop, $l_1$, handles the choice of training and validation samples. The two middle nested loops, $l_2$ and $l_3$, perform the LM optimization over the selected training set with $n_v$-fold cross-validation. The inner most loop, $l_4$, is to modify the parameter $\lambda$ using equation~\eqref{eq:LM}.

Assume that the network is properly initialized and the dataset has $n_D$ samples in it, each is an input/output trajectory segment with T time steps:
\begin{equation}
	D=\big{\{}(\mathbf{u}_d,\mathbf{z}_d)\big{\}},|D|=n_D,
\end{equation}
where $|.|$ states the cardinality of a set and
\begin{equation}
	\mathbf{u}_d=\begin{bmatrix}u_d(1) & \dots u_d(T)\end{bmatrix},\mathbf{z}_d=\begin{bmatrix}z_d(1) & \dots z_d(T)\end{bmatrix}.
\end{equation}
At each iteration of $l_1$ we pick $n_S=n_{tr}+n_v$ samples from $D$ to form the subset $D_S$. 
\begin{equation}
	D_S\subset D, |D_S|=n_s
	\label{eq:DS}
\end{equation}
The number of trajectory segments that we simultaneously train the network on is $n_{tr}$, and $n_v$ is the number of trajectory segments that we validate the network on. Both constants should be kept small to reduce the computational complexity of the training process.  Then, at each iteration of $l_2$, we divide $D_S$ into two sets: the training set, $D_{tr}$, and the validation set, $D_v$, so that $|D_{tr}|=n_{tr}$ and $|D_v|=n_v$.

Having the training and validation sets, the optimization starts in loop $l_3$ and continues until the validation fails, i.e. the error over validation set starts to increase. At each iteration of $l_3$, we run the network over the entire $D_{tr}$, update and collect the network Jacobians, as described in Section~\ref{sec:deriv}, and calculate the errors. Then we form the Jacobian in equation~\eqref{eq:LM} as described in equation~\eqref{eq:LMJ}. The weight update process is performed in loop $l_4$, where we update the network weights using equation~\eqref{eq:LM} starting with an initial (usually small) $\lambda_{0}$. At each iteration of $l_3$, we check if the error is decreased/increased and update $\lambda$ accordingly. The detailed steps of this algorithm are defined in Algorithm~\ref{alg:LM}. Although specific choices are made in terms of stopping criteria, step size, etc., these choices are not a requirement of the algorithm and may easily be modified for application of the MODERNN algorithm to other problem instances.

It is important to note that each weight update is performed using the Jacobian that is computed over a number of flights at once.  We do not apply each training sample inside $D_{tr}$ individually, but instead carry out the training simultaneously over the entire set, $D_{tr}$. The function Minibatch is so named to refer to the training over the entire batch of trajectory segments, $D_{tr}$.
\begin{algorithm}
	\caption{L Learning Algorithm for MODERNN}\label{alg:LM}
	\begin{algorithmic}[1]
		\For {$l_1$=1 to $\Big{\lfloor}\frac{n_D}{n_S}\Big{\rfloor}$}\Comment{loop $l_1$}
			\For {$l_2$=1 to $|S|$}\Comment{loop $l_2$}
				\State $D_{tr}\gets D_S\big{(}S(l_2)\big{)}$
				\State $D_{v}\gets D_S\backslash D_{tr}$
				\State $\gamma_1\gets 1$
				\State $\lambda\gets\lambda_{0}$
				\While {$\gamma_1$}\Comment{loop $l_3$}
					\State $\mathbf{e}_{v,0}\gets \Call{Minibatch}{D_{v},\mathbf{p},0}$
					\State $[\mathbf{e}_{tr,0},\mathbf{J}]\gets \Call{Minibatch}{D_{tr},\mathbf{p},1}		$
					\State $\gamma_2\gets 1$
					\While{$\gamma_2$}\Comment{loop $l_4$}
						\State $\Delta\mathbf{p}\gets -(\mathbf{J}^T\mathbf{J}+\lambda\mathbf{I})^{-1}\mathbf{J}^T\mathbf{e}$
						\State $\mathbf{e}_{tr,1}\gets \Call{Minibatch}{D_{tr},\mathbf{p}+\Delta\mathbf{p},0}$
						\If{$\mathbf{e}_{tr,1}^T\mathbf{e}_{tr,1}<\mathbf{e}_{tr,0}^T\mathbf{e}_{tr,0}$}
							\State $\lambda\gets\lambda\times\frac{2}{3}$
							\State $\gamma_2\gets 0$
							\State $\mathbf{p}\gets\mathbf{p}+\Delta\mathbf{p}$
						\Else
							\State $\lambda\gets\lambda\times\frac{3}{2}$
							\If{$\lambda>\lambda_{Max}$}
								\State $\gamma_2\gets 0$
							\EndIf
						\EndIf
					\EndWhile
					\State $\mathbf{e}_{v,1}\gets \Call{Minibatch}{D_{v},\mathbf{p},0}	$
					\If{$\mathbf{e}_{v,1}^T\mathbf{e}_{v,1}>\mathbf{e}_{v,0}^T\mathbf{e}_{v,0}$}\Comment{Validation fails.}
						\State $\gamma_1\gets 0$
					\EndIf
				\EndWhile
			\EndFor
		\EndFor
		\Function{$[\mathbf{e},\mathbf{J}]$=Minibatch}{$D_{tr},\mathbf{p},\delta$}
			\For {$s=1$ to $|D_{tr}|$}
				\State $\mathbf{y}\gets\Omega_L(\mathbf{u}_s)$
				\State $\mathbf{e}_s\gets[e(1)\hspace{5pt}\dots\hspace{5pt}e(T)]^T$\Comment{eq.~\eqref{eq:ey}}
				\If{$\delta=1$}
					\State $\mathbf{J}_s\gets\frac{\partial\mathbf{e}_s}{\partial\mathbf{p}}$\Comment{eq.~\eqref{eq:LMJ}}
				\EndIf
			\EndFor
				\State $\mathbf{e}^T\gets\big{[}\mathbf{e}_1^T\hspace{5pt}\dots\hspace{5pt}\mathbf{e}_{|D_{tr}|}^T\big{]}$
			\If{$\delta=1$}
				\State $\mathbf{J}^T\gets\big{[}\mathbf{J}_1^T\hspace{5pt}\dots\hspace{5pt}\mathbf{J}^T_{|D_{tr}|}\big{]}$ 
			\EndIf
		\EndFunction		
	\end{algorithmic}
\end{algorithm}
The formulation of MODERNN and the presented basic algorithm is quite flexible both in terms of attainable architectures and applicable training methods, as the network Jacobians can be computed explicitly. In the version of MODERNN with all inter-layer connections, the problem of vanishing/exploding gradient is less severe because direct connections from the output layer(s) to all other layers facilitate the error information transfer between layers. For function approximation problems, the only information that the network receives from outside the network is the Sum-of-Squared Errors.  In traditional multilayer networks, the error information starts to flow from the output layer through the middle layers back to the input layer in a sequential manner. At each layer this information is either attenuated or amplified, and because network weights are usually initialized in the range $[-1,1]$, attenuation occurs. In the MODERNN, the error information travels through a number of different paths, this multipath transfer of information contributes to faster and better learning and less attenuation of the error gradient in former layers, as presented in Section~\ref{sec:sim}. 
 
\section{Simulation Results}
\label{sec:sim}
\begin{figure}[h]
	\vspace{-12pt}
	\begin{centering}
		\includegraphics[scale=0.44]{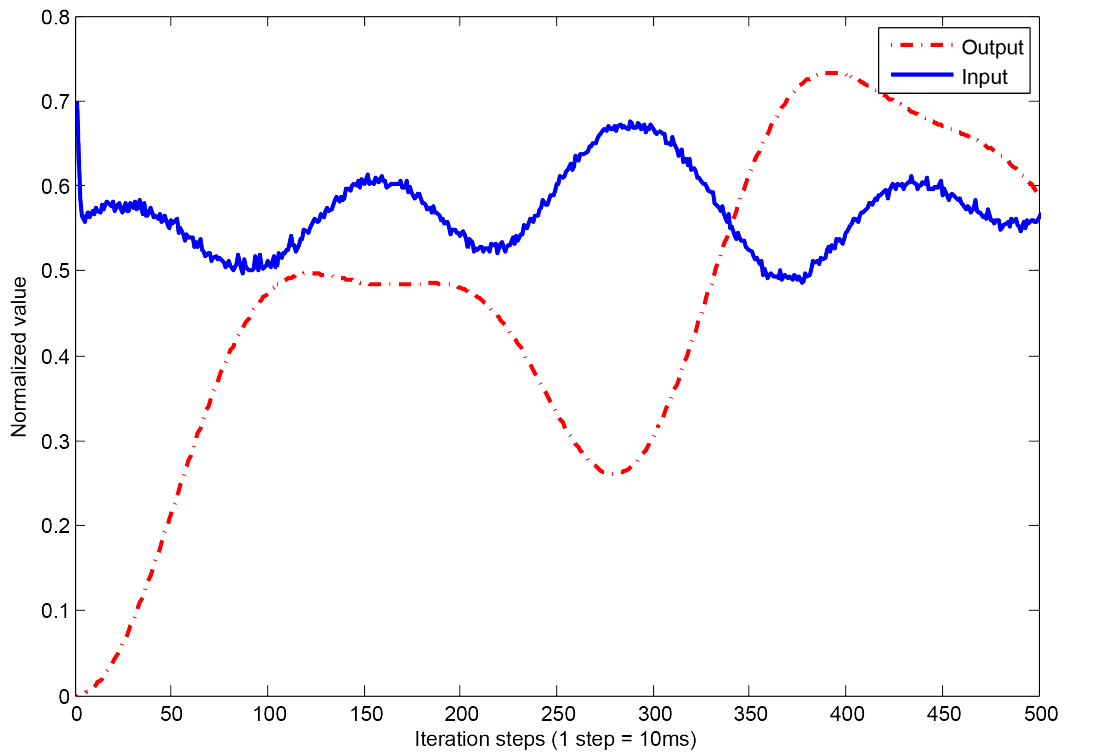}
		\caption{ A generated data sample.}
		\label{fig:DATA}
	\end{centering}
	\vspace{-18pt}
\end{figure}

We apply the MODERNN structure to the problem of learning quadrotor altitude dynamics.  The details of quadrotor modelling are well established and can be found in, among others,~\cite{hoffmann2011}. For our simulation, the model to generate the altitude data is given by
\vspace{-5pt}
\begin{equation}
	\begin{aligned}
		\ddot{z} & = \frac{1}{m}\bigg{(}k_tu^2\big{(}1+f_{ge}^2\big{)}-c_d\dot{z}-mg+\eta\bigg{)}\\
		f_{ge} & =k_{ge}\frac{h_{ge}-\min(h_{ge},z)}{h_{ge}}.
	\end{aligned}
	\label{eq:QMODEL}
\end{equation}
In equation~\eqref{eq:QMODEL}, $z$ is the altitude of the vehicle and $u$ is the sum of all four motor speeds (in RPM), which relates directly to the thrust produced. The mass of the quadrotor is $m$, $k_t$ is the thrust co-efficient, $c_d$ is the drag coefficient and $f_{ge}$ is a simple model for ground effect. The ground effect acts at altitudes lower than $h_{ge}$, and $k_{ge}$ is the ground effect co-efficient. Finally, $\eta$ is a white noise. Table~\ref{tab:PARAMS} lists the values used for data generation. To generate a more realistic dataset, we would like to have flyable trajectories, and as such, a random noise as the input to equation~\eqref{eq:QMODEL} is not considered.  Instead, each altitude trajectory was created by applying a random sum of $10$ sinusoids with variable frequencies in range of $[1,10]Hz$ as input. To capture the ground effect, altitude is varied in the range $[0,2]$ meters over the whole dataset.  The dataset is collected using a $f_s = 10Hz$ sampling frequency. In Fig.~\ref{fig:DATA} one sample of the dataset with normalized values is illustrated.
\begin{table}
\renewcommand{\arraystretch}{2}
	\begin{center}
		\setlength{\tabcolsep}{4.5pt}.
		\begin{tabular}{|c|c|c|c|c|}
			\hline
			$m$ & $k_{ge}$ & $h_{ge}$ & $k_t$  & $z_{max}$\\
			\hline
			$1$ kg & $0.5$ & $1$ m & $1.95\times 10^{-5}\frac{\mbox{Ns}^2}{\mbox{rad}}$& $2$ m\\
			\hline
		\end{tabular}
	\end{center}
	\caption{Quadrotor parameters used in our simulations}
	\label{tab:PARAMS}
	\vspace{-20pt}
\end{table}

Having generated the dataset, the MODERNN, RMLP and NARX architectures are trained on it. Our experiments on series-parallel learning of NARX using the MATLAB toolbox failed when we closed the loop for prediction. Instead, we implemented our own parallel-model NARX, calculated the analytic derivatives of the network and implemented a training process based on the same method we presented in Section~\ref{sec:deriv}. We validated our implementation on examples given in~\cite{Narendra90}. It is therefore important to highlight that we do not use any form of teacher forcing in training the three networks; they are trained and evaluated in a closed loop fashion only.

The parameters under investigation are: the number of layers $L$, the number of hidden neurons in each layer $h$, and  the size of the minibatch $n_{tr}$. For NARX, the number of delays over input and output $n_d$ is also investigated. Note that in multilayer cases, we choose to have the same number of neurons in all layers.  Also, remember that the last layer has a linear output as is always the case for function approximation applications. The result are summarized in Table~\ref{tab:RESULTS1}. In this table, $E_m$ is the mean value of the errors over the entire dataset and $t_{tr}$ is the training time on an i7 Core machine. In general, we observe that finding a working configuration becomes increasingly harder from MODERNN to RMLP and to NARX. It was not possible to find a working architecture for NARX with $n_{tr}=5,10,15$ and RMLP with $n_{tr}=5,10$, while MODERNN can learn on $n_{tr}=5$ and higher. For each of the reported cases we run the training 5 times, each time with a different weight (random) initialization. The best results are reported only. As a result of the ability to accurately learn the quadrotor model with a MODERNN with 48 weights, computation times were significantly improved over both RMLP and NARX, to approximately 0.5 hours from 5.5 and 7.5, respectively.\\
\begin{figure}[h]
	\vspace{-5pt}
	\includegraphics[scale=0.5]{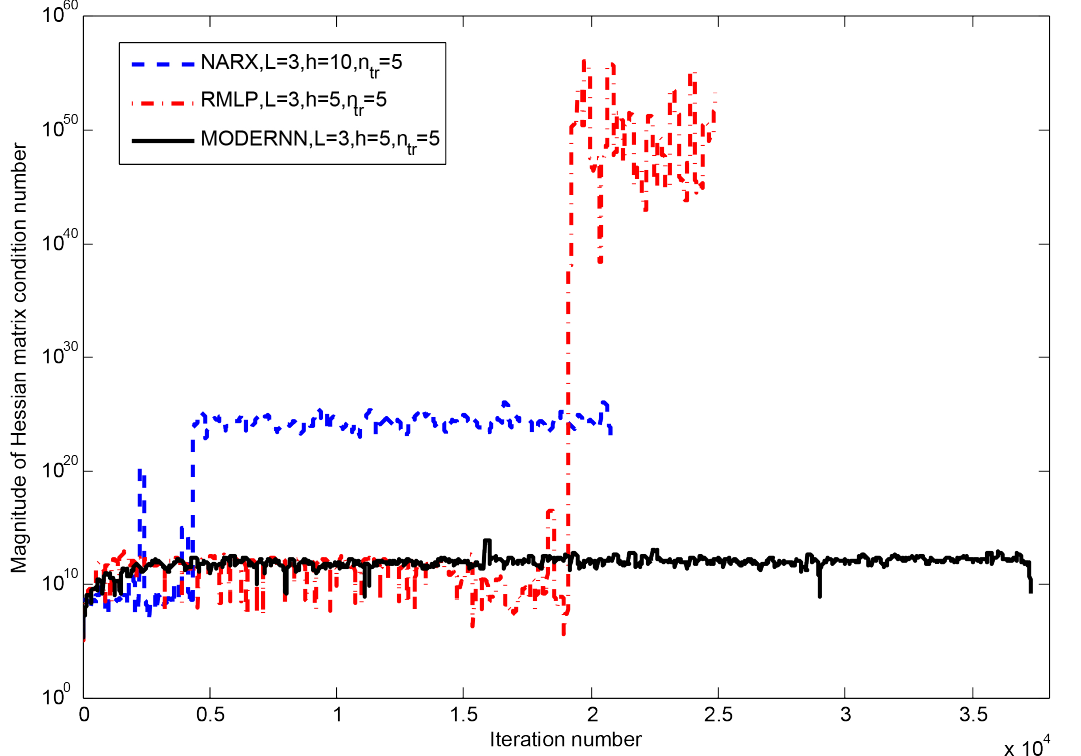}
	\caption{ Comparison between Hessian condition numbers.}
	\label{fig:COND}
	\vspace{-15pt}
\end{figure}
\begin{table}
\renewcommand{\arraystretch}{1}
	\begin{center}
		\setlength{\tabcolsep}{4.5pt}.
		\begin{tabular}{|c||c|c|c|c|c||c|c|}
			\hline
			Network & $L$ & $h$ & $q$ & $n_{tr}$ & $n_d$ & $E_m$ & $t_{tr}(h)$ \\
			\hline
			\multirow{6}{*}{MODERNN} & 2 & 5 & 48   & 5 & NA &0.424 & $\approx$0.5\\ 
			 & 3 & 5 & 195 & 5 & NA &0.322 & 1\\ 
			 & 2 & 5 & 48   & 10 & NA &0.152 & 1.7\\ 
			 & 3 & 5 & 195 & 10 & NA &0.612 & 1.2\\ 
			 & 2 & 5 & 48   & 20 & NA &0.197 & 2.2\\ 
			 & 3 & 5 & 195 & 20 & NA &0.088 & 3.5\\ 
			\hline
			\multirow{6}{*}{RMLP} & 2 & $5,10,20$ & $42,132,462$ &5,10 &NA & - & - \\
			 & 3 & $5,10,20$ & $97,342,1282$ & 5,10 &NA & - & - \\
			 & 2 & $10$ & $132$ &15 &NA & 1.01 & 2.1 \\
			 & 3 & $5$ & $97$ & 15 &NA & 0.877 & 2.9 \\
			 & 2 & $5$ & $42$ & 20 &NA & 0.202 & 5.5 \\
			 & 3 & $5$ & $97$ & 20 &NA & 0.122 & 9.3 \\
			\hline
			\multirow{6}{*}{NARX} & 2 & $5,10,20$ & $56,111,221$ & 5,10,15 & 4 & - & - \\
			 & 2 & $5,10,20$ & $66,131,261$ & 5,10,15 & 5 & - & - \\
			 & 3 & $5,10,20$ & $86,221,641$ & 5,10,15 & 4 & - & - \\
			 & 3 & $5,10,20$ & $96,241,681$ & 5,10,15 & 5 & - & - \\
			 & 2 & $10$ 	 & $111$ 	& 20 & 4 & 0.7175 & 6 \\
			 & 3 & $10$ 	 & $221$ 	& 20 & 6 & 0.332 & 7.5 \\
			\hline
		\end{tabular}
	\end{center}
	\caption{Results of experiments of modelling a simulated quadrotor altitude using MODERNN,RMLP and NARX.}
	\label{tab:RESULTS1}
	\vspace{-25pt}
\end{table}

One of the main reasons for failure in optimization problems is that the magnitude of the condition number of the Hessian becomes extremely large, and accuracy is lost due to the limits of machine precision. To have a picture of how the condition number behaves typically in the learning process for each of the three networks, in Fig.~\ref{fig:COND} a window-averaged (smoothed) version is plotted versus iteration steps. We see that for the same $n_{tr}$, NARX and RMLP eventually lead to condition numbers much larger than MODERNN as the optimization progresses and therefore introduce more numerical error in the solution. This may contribute to failure of NARX and RMLP to learn on small batches. Since the only difference between MODERNN and RMLP is the presence of forward connections in MODERNN, both having the same number of layers and hidden neurons, we conclude that forward connections contribute to a less ill-conditioned Hessian.
\vspace{-5pt}
\section{Conclusion}
\label{sec:conc}
In this paper, we present a modular and flexible class of RNN architectures for modelling of high-order dynamic and nonlinear systems. MODERNN not only encapsulates a large class of already well-studied RNNs, but also provides a modular and easy-to-implement way of calculating network Jacobians. We also demonstrate that feedforward connections in a RNN can play a significant role in alleviating vanishing/exploding gradient in space. We  provide a basic learning algorithm based on the LM optimization method, which illustrated successful application in learning altitude dynamic of a simulated quadrotor. By enabling smaller networks to learn the same complexity dynamics, we have dramatically reduced the necessary computation time to achieve reliable models for dynamical systems of moderate complexity with significant nonlinearities present.
\vspace{-10pt}
\bibliographystyle{IEEEtran}
\bibliography{NNBIB}

\end{document}